# Siamese Meta-Learning and Algorithm Selection with 'Algorithm-Performance Personas' [Proposal]

**Joeran BEEL,** Trinity College Dublin, School of Computer Science, ADAPT Centre, Ireland & The University of Siegen, Department of Electrical Engineering and Computer Science, Germany.

**Bryan TYRELL,** Trinity College Dublin, School of Computer Science, Ireland

**Edward BERGMAN,** Trinity College Dublin, School of Computer Science, ADAPT Centre, Ireland & The University of Siegen, Department of Electrical Engineering and Computer Science, Germany.

**Andrew COLLINS,** Trinity College Dublin, School of Computer Science, ADAPT Centre, Ireland

**Shahad NAGOOR,** Trinity College Dublin, School of Computer Science, ADAPT Centre, Ireland

**Abstract.** Automated per-instance algorithm selection often outperforms single learners. Key to algorithm selection via meta-learning is often the (meta) features, which sometimes though do not provide enough information to train a meta-learner effectively. We propose a Siamese Neural Network architecture for automated algorithm selection that focuses more on 'alike performing ' instances than meta features. Our work includes a novel performance metric and method for selecting training samples. We introduce further the concept of 'Algorithm Performance Personas' that describe instances for which the single algorithms perform alike. The concept of 'alike performing algorithms ' as ground truth for selecting training samples is novel and provides a huge potential as we believe. In this proposal, we outline our ideas in detail and provide the first evidence that our proposed metric is better suitable for training sample selection that standard performance metrics such as absolute errors.

## 1 INTRODUCTION

### 1.1 Background

The algorithm selection problem is a long-time recognized problem that describes the challenge of identifying the ideal algorithm, and configuration, for a given computational task [1], [2]. Algorithm selection may be done "per-set" [3], [4] to identify the best algorithm for an entire dataset or application (e.g. a news website). This type of algorithm selection is sometimes also called "macro" algorithm selection [5]. In contrast, "per-instance" [3], [4] or "micro" [5] algorithm selection aims at identifying the best algorithm for every single instance in a dataset or for a sub-set of instances (e.g. one algorithm per user or user group [6]).

The per-set algorithm selection problem is demonstrated by one of our experiments in the domain of recommender systems [7]. We experimented with five recommendation algorithms on six news websites, and on almost every website a different algorithm performed best and worst (Figure 1). The 'most popular' algorithm performed best on *cio.de* and *tecchannel.de* but worst on *ksta.de* and *tagesspiegel.de*; content-based filtering performed best on *motor-talk.de* but worst on *tecchannel.de*; 'most popular sequence' performed best on *sport1.de* but second worst on *cio.de* and *tecchannel.de*. Given this information, the developer of a new website would hardly know, which algorithm to choose for the new website and hence face the (per set) algorithm selection problem.

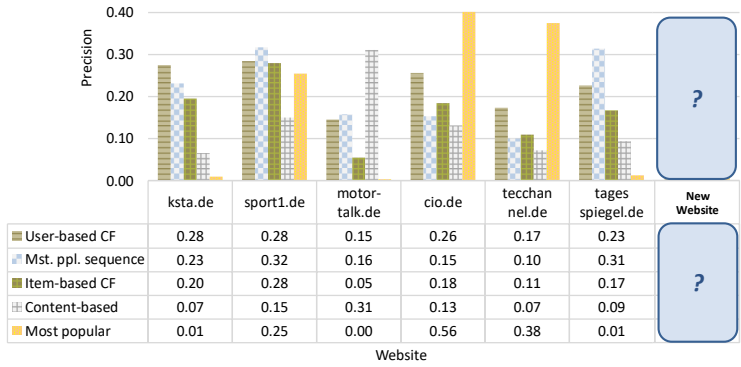

**Figure 1: Performance (Precision) of five recommendation algorithms on six news websites** [7]

The per-instance algorithm selection problem is also demonstrated by our previous experiments [7]. As mentioned, on *sport1.de,* the 'most popular sequence' algorithm performs best overall, i.e. on set-level (Figure 1). However, looking at the data in more detail, reveals a different picture (Figure 2). The algorithm only performs best between 18:00 and 4:00 o'clock. The remaining time, the 'Most Popular' algorithm performs best (Figure 2), which is only fourth-best overall (Figure 1). Consequently, the most effective recommender system would be a system that used the two algorithms depending on the time of the day. However, time of the day is only one factor. There is typically a plethora of factors that affect algorithm performance. In our experiment this was, among others, age, gender, and the number of displayed recommendations [7]. This makes the goal of identifying the best algorithm for a given instance, or sub-set of instances, complex.

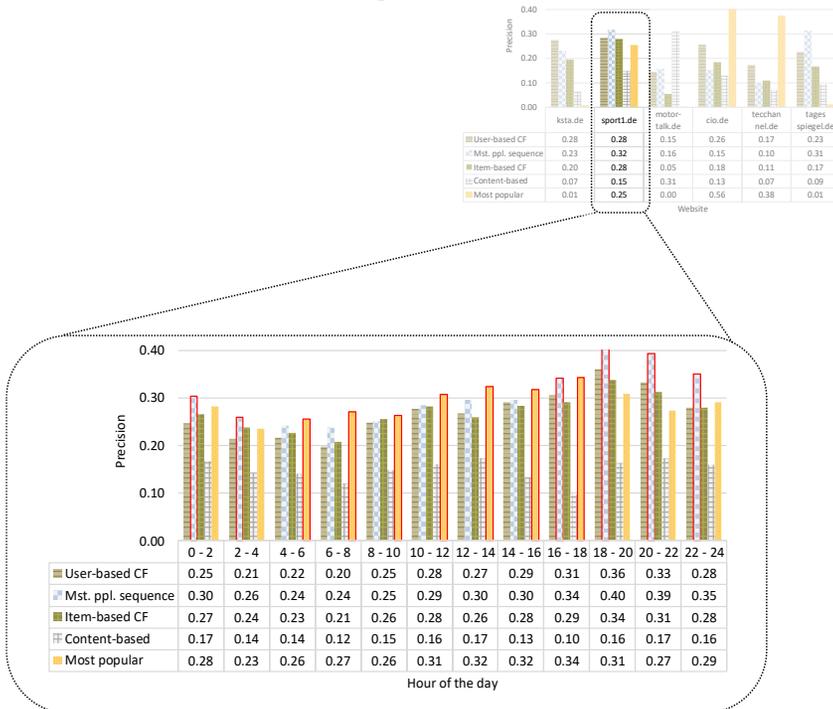

**Figure 2: Performance of the five algorithms on sport1.de, based on time of the day**

The solution to the algorithm selection problem on both set and instance level is *automated algorithm selection*. Automated algorithm selection refers to either optimization techniques or meta-learning ('learn to learn') [8]. In our work, we focus on meta-learning. Meta-learning in its core is a classification task. Based on the (meta-) features of the instances or datasets and known performance measures of algorithms on these data points, a machine learning model is trained to predict, which algorithm will perform best on a data point[1]. The meta-learner can then predict for any new unlabeled instance or dataset the potentially best performing algorithm. Meta-learning can also be implemented as regression model, in which the performance (e.g. RMSE) of each algorithm is predicted. The algorithm with the best predicted performance is then applied to the dataset or instance. In case of per-instance algorithm selection both, meta-learner and single algorithms may use the same features for training -- only the target differs. This is illustrated in Figure 3. The ordinary pool of single machine learning algorithms $A = \{a_1...a_n\}$ aims to predict target $t$. Performance of the algorithms is measured e.g. by the error rate of the predictions. A meta-learner $a_{ml}$ is trained on the same features $x_1...x_m$ but with the performance measures (e.g. error rates) as the target. The algorithm with the best predicted performance would then be applied to predict the target.

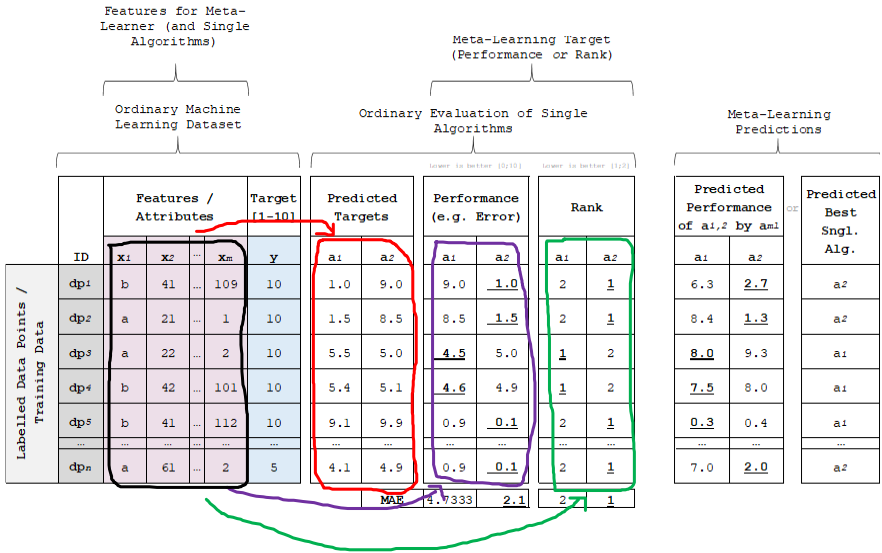

**Figure 3: Illustration of the difference between 'ordinary' (red) machine learning and meta-learning (green/purple). The input features may be the same, but the target differs.**

For set-level algorithm selection, new meta-features typically are needed that describe the dataset. In automated machine learning (AutoML), meta-learning is often used to warm-start the algorithm selection process [8], [9]. First, a meta-learner predicts a small set of potentially well performing algorithms, and these algorithms are then further optimized to find the best performing algorithm and configuration. It is important to note that in set-level algorithm selection the goal is to identify the best single algorithm for the given set. In per-instance algorithm selection, the goal is to create some kind of switching ensemble that uses a number of single algorithms to eventually achieve a higher performance compared to using only the single best algorithm. It is further important to note that per-instance algorithm selection differs from typical ensemble methods. Ensemble methods like stacking require that all algorithms are somewhat similar, e.g. that they all are regression (or classification) algorithms that make a numeric prediction (or classification). With per-instance algorithm selection, it is possible to select completely different algorithms for different instances. This is valuable e.g. for recommender systems where e.g. collaborative filtering, content-based filtering, most popular or stereotype recommendations could be used for different instances.

---

[1] The meta learner may also predict a number of potentially well performing algorithms but for sake of clarity we focus on a single algorithm.

There is an active research community on automated algorithm selection with dedicated workshops and competitions [10–18], particularly in the fields of automated machine learning (AutoML), meta-learning, (hyperparameter) optimization, and neural architecture search. Automated algorithm selection algorithms have shown to be effective in many disciplines including databases [19], [20], SAT [21], machine learning [17], [22–26], data mining [27], information retrieval [11], [28–32], cloud resource allocation [33], game solvers, electronic design [34] and material sciences [35], recommender systems [6], [36–43] and reference-string parsing [44], [45].

## 1.2   Research Problem

The algorithm-selection community focuses to a large extent on the creation of (meta-) features [46–49]. This is no surprise as (meta-) features are fundamentally important to predict algorithm performance. However, at least sometimes, features are not suitable to (directly) predict algorithm performance. We illustrate this problem in Figure 4, a made-up example that we will use throughout this manuscript. The figure illustrates a typical dataset with $n$ datapoints $dp_i$. Each $dp_i$ has $m$ features and a target $t$ (in case of a recommender system e.g. the rating of a product). The illustration shows the predicted targets of two algorithms $a_1$ and $a_2$, their prediction errors /target-predicted target/ and the rank (1= best performing algorithm on that instance). The data points $dp_1$ and $dp_2$ are not similar in terms of features (red), but the algorithms $a_1$ and $a_2$ perform alike on them (green).

|  | Features / Attributes | | | Target [1-10] | Predicted Targets | | Performance (e.g. Error) | | Rank | |
|---|---|---|---|---|---|---|---|---|---|---|
| ID | $x_1$ | $x_2$ | $x_m$ | y | $a_1$ | $a_2$ | $a_1$ | $a_2$ | $a_1$ | $a_2$ |
| $dp_1$ | b | 41 | 109 | 10 | 1.0 | 9.0 | 9.0 | **1.0** | 2 | **1** |
| $dp_2$ | a | 21 | 1 | 10 | 1.5 | 8.5 | 8.5 | **1.5** | 2 | **1** |
| $dp_3$ | a | 22 | 2 | 10 | 5.5 | 5.0 | **4.5** | 5.0 | **1** | 2 |
| $dp_4$ | b | 42 | 101 | 10 | 5.4 | 5.1 | **4.6** | 4.9 | **1** | 2 |
| $dp_5$ | b | 41 | 112 | 10 | 9.1 | 9.9 | 0.9 | **0.1** | 2 | **1** |
| ... | | | | | | | | | | |
| $dp_n$ | a | 61 | 2 | 5 | 4.1 | 4.9 | 0.9 | **0.1** | 2 | **1** |
| $dp_{t1}$ | b | 45 | 105 | ??? | ??? | ??? | ??? | ??? | ??? | ??? |
| ... | | | | | | | | | | |
| $dp_{tp}$ | a | 22 | 1 | ??? | ??? | ??? | ??? | ??? | ??? | ??? |

**Figure 4: Illustration of a machine learning dataset and performance of two algorithms a1 and a2**

By "perform alike" we mean that the same algorithm ($a_2$) performs best, and prediction errors of the two algorithms are also similar ($a_1$ has errors of 9 and 8.5 on dp1 and dp2 respectively and $a_2$ has errors of 1 and 1.5 respectively). In contrast, $dp_4$ and $dp_5$ have similar features (purple), but the algorithms perform not alike on them (orange): for $dp_4$, algorithm $a_1$ performs best (lowest error), but on $dp_5$, algorithm $a_2$ performs best. Training a standard machine learning model on these features to predict the performance for unseen instances (blue), or applying a kNN approach, would likely not lead to promising results.

## 1.3 Research Goal

While traditional meta learning approaches go from the features to the targets, we aim to go the opposite way. We aim to leverage the knowledge that two instances are (not) similar given their algorithm performance, regardless of their features. In other words, our goal is to make a meta-learner learn that two data points are similar if algorithms perform alike on these two data points. For a new data point -- for which it is not known, which algorithm performs best -- the network shall then be able to identify the most similar datapoints in the set of data points for which the best performing algorithms are known. The algorithm that performs best on those similar data points likely will perform best on the new data point. This means based on performance measures $P$, the network shall learn to what extent two data points, which are described by their feature vector $x$, are similar. Hence, the goal is to learn a transformation that transforms a data point's features $x$ into an embedding $x'$. Data points being similar in terms of performance should also be similar in terms of embeddings. The goal is illustrated in Figure 5. As explained earlier, algorithms $a_1$ and $a_2$ perform similar on the data points $dp_1$ and $dp_2$. Hence, by definition, $dp_1$ and $dp_2$ are similar, although their features may (or may not) differ. Based on this defined ground truth, we want the network to learn to transform the features of $dp_1$ and $dp_2$ into the embeddings $dp_{1'}$ and $dp_{2'}$ so that the two data points $dp_{1'}$ and $dp_{2'}$ are close to each other in the embedding space.

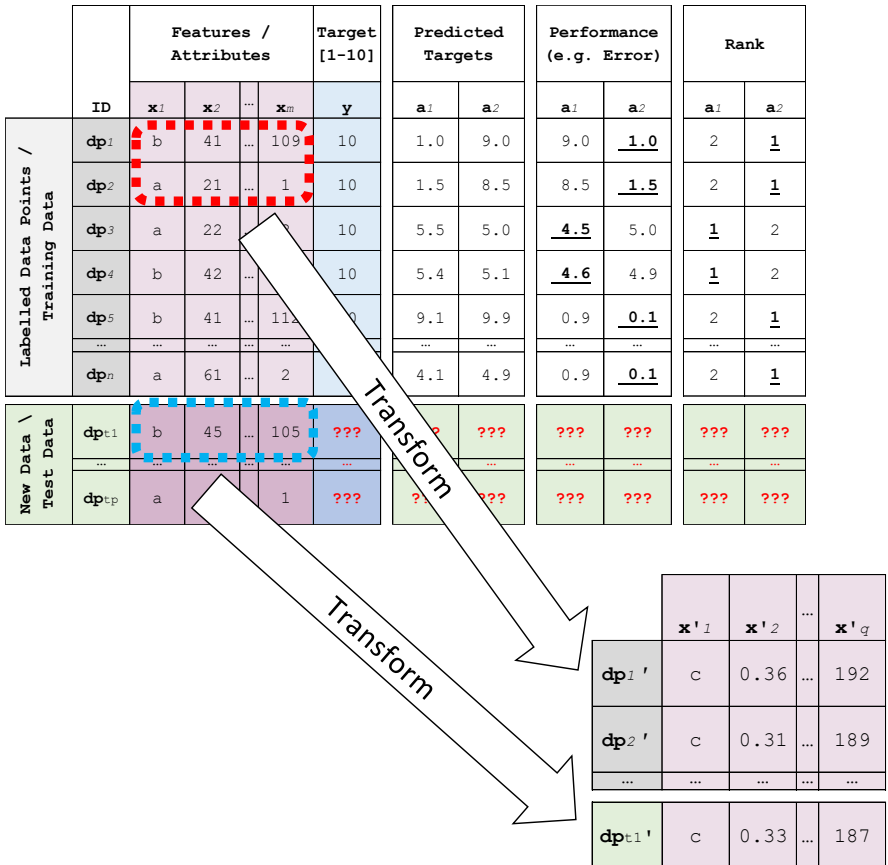

**Figure 5: The goal is to learn an embedding to represent performance similarity. While the algorithms perform alike on dp1 and dp2, the data points' features differ. The network learns to transform the features into embeddings that are similar.**

### 1.4   Our Contribution

Our contribution is threefold.
1. We propose a novel performance metric, or normalization scheme, to express the performance of algorithms on data points.
2. We propose the concept of 'Algorithm Performance Personas' (APP) to generically describe data points that perform alike.
3. We propose a Siamese Neural Network architecture to predict alike performing instances. The Siamese Neural Network learns the similarity of instances based on the algorithms' performances on these instances. This includes novel strategies to select positive and negative training samples, as well as simple and difficult to learn samples.

Please note that we are currently only presenting the proposal and some initial evidence that this proposed architecture may be promising. More work will be needed to implement the architecture and evaluate it – first results will be presented in an upcoming publication [50].

Our work builds upon the ideas of Pulatov & Kotthoff [25] as well as Kim et al. [9]. Pulatov & Kotthoff brought forward the idea that an algorithm is more than just a featureless class that is to be predicted. Algorithms have characteristics that affect how they perform on a dataset or instance. Consequently, Pulatov & Kotthoff included algorithm features in the training of their meta learners. While we do not use algorithm features for our approach, Pulatov & Kotthoff's work inspired us to leverage information beyond standard meta-features. In our case, we leverage the 'performance similarity' of algorithms on the instances to identify alike-performing instances that are suitable for training a Siamese Neural Network. Kim et al. [9] used Siamese Neural Networks for algorithm selection before. However, they did not use algorithm performance to select training instances but the similarity of (meta)-features. We believe that our approach aligns better with the original idea behind Siamese Neural Networks.

## 2   BACKGROUND: SIAMESE NEURAL NETWORKS & FEW-SHORT IMAGE CLASSIFICATION

Siamese Neural Network architectures [51] have become highly successful for few-shot and metric learning, particularly for image classification [52], [53] but also other applications. The goal of few-shot image classification is to classify images for which only few labelled samples exist – too few to train a model in the traditional way. Figure 6 illustrates this concept. The labelled dataset contains images of the two persons 'Arnold Schwarzenegger' (photos 1 and 2) and 'Joeran Beel' (photos 3, 4, and 5). There could be hundreds of classes more, each with only a few labelled images. A traditional (deep) machine learning algorithm (e.g. a CNN) would have difficulties in learning the classes with just few samples per class. In the example, the unlabeled image 6 likely would be misclassified since its features (pixels, color space, etc.) are more similar with photos of the class 'Arnold Schwarzenegger' than the correct class 'Joeran Beel'.

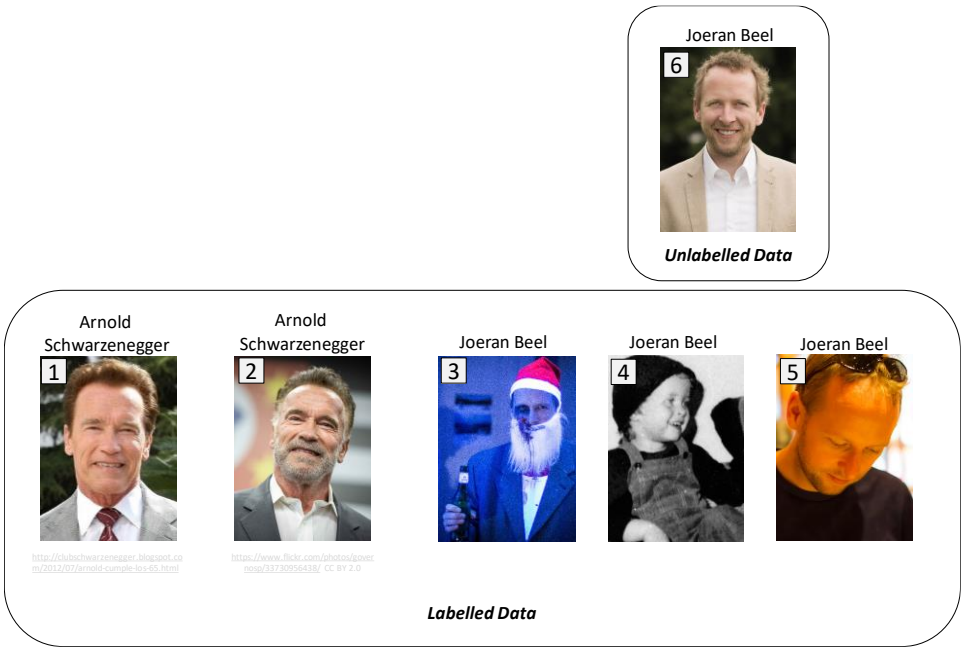

**Figure 6: Illustration of a few-shot image classification dataset with only few images for each class.**

To address this challenge, Siamese Neural Networks learn to calculate the similarity between any two data points, instead of learning to predict a class directly. This learned similarity function can be applied to any two datapoints including one unlabeled and one labelled data-point. If the distance between the unlabeled data point and the known data point is smaller than a margin $\alpha$, the unabelled data point is assumed to be of the same class as the labelled data point. In the example, the Network would learn that images 1 and 2 are similar/identical and that images 3, 4 and 5 are similar – even if their actual features (color etc.) differ. Such a similarity function is more likely able to predict that the unlabeled image 6 is similar to image 3, 4, and 5 and consequently assign the correct class to image 6.

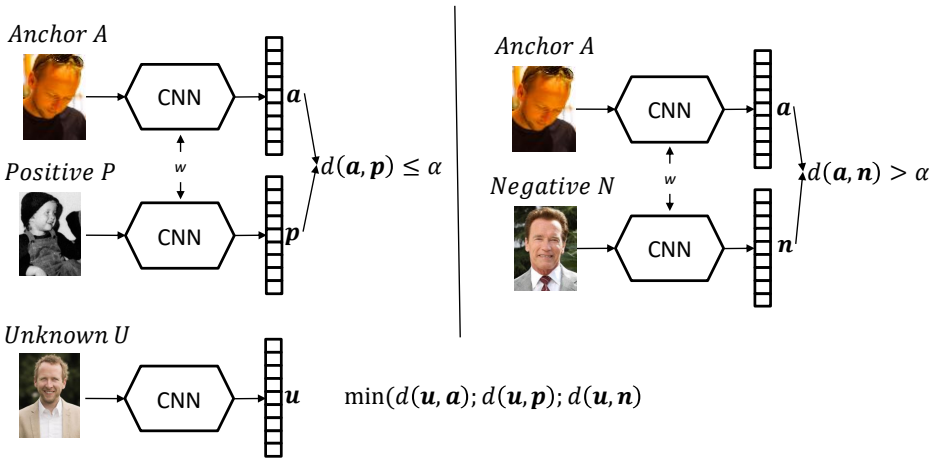

**Figure 7: Siamese Neural Network Illustration for Image Classification**

A Siamese Neural Network architecture consists of any two neural networks (e.g. CNNs). One network receives an 'anchor' image and a 'positive' training sample as input (illustrated in Figure 7). The other network receives the same anchor image and a negative sample as input. The two networks are then trained to output embeddings that are close to each other ($<$ margin $\alpha$) in the embedding space for the positive example, and far from each other ($> \alpha$) for the negative example. For new unlabelled instances, the network transforms the image features into an embedding, and predicts the class based on the class of those image(s) being closest to the input image in the embedding space. There are multiple variations of Siamese Neural Networks but most important for our work is the core idea.

## 3 SIAMESE ALGORITHM SELECTION & ALGORITHM PERFORMANCE PERSONAS

### 3.1 General Idea

Analog to Siamese Neural Networks for image classification, we propose a Siamese Neural Network architecture as illustrated in Figure 8 and similar to Kim et al. [9]. Our architecture comprises of two identical neural networks (MLPs in our case). Network 1 receives a data point (anchor) $dp_A$ as input, whereas the datapoint is represented by its features. Network 2 receives another data point as input, the positive sample $dp_P$, whereas $dp_P$ is similar to $dp_A$. Similarity is not based on (meta) features as Kim et al. [9] do, but based on performance of algorithms on these datapoints. This means that the single algorithms in the pool all perform similar on both $dp_P$ and $dp_A$ (how similarity is calculated exactly is explained later). The network is then trained to create the embeddings $a$ and $p$ that are close to each other in the embedding space, whereas 'close' means less than a margin α (how distance is measured exactly is explained later). In addition, data points for which algorithms do not perform similar to the anchor data point, are used as negative training examples $dp_N$. The network is trained so that the embeddings $a$ and $n$ are not close to each other in the embedding space, whereas 'not close' means larger than margin α. A new unlabeled data point $dp_U$ is then transformed into an embedding $u$. The system finds the k nearest neighbors to $u$ in the embedding space with a distance less than α. The algorithm that performs best for the k nearest neighbors is then chosen to be applied on $dp_U$.

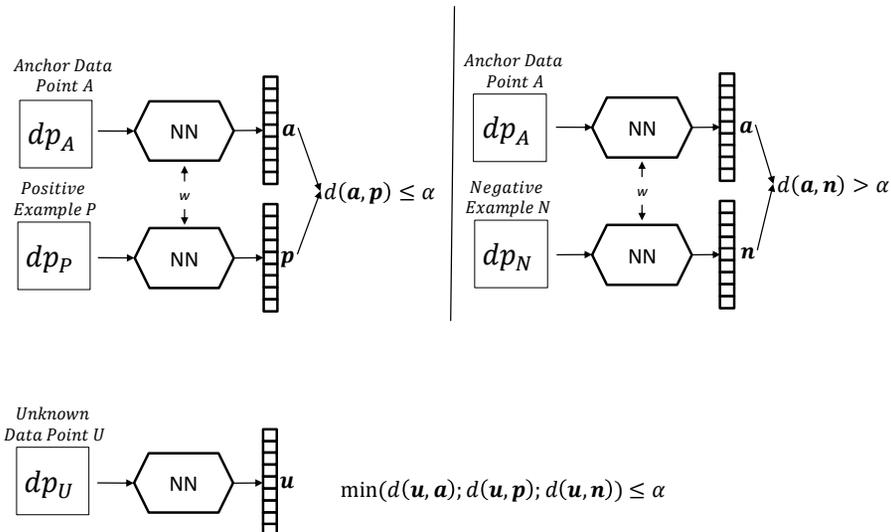

Figure 8: Siamese Neural Network Architecture for Algorithm Selection.

Figure 9 further illustrates the concept with our example. As mentioned, $dp_1$ and $dp_2$ are similar based on their performance. Hence, one of the two data points becomes the anchor point, and the other one the positive sample. $dp_4$ is selected as negative sample as algorithms perform different on $dp_4$ than on $dp_1$ or $dp_2$. The network then learns to transform $dp_1$ and $dp_2$ into embeddings $dp_{1'}$ and $dp_{2'}$ being close to each other in the embedding space, whereas $dp_{4'}$ would be far in the embedding space. The unlabeled data point $dp_{t1}$ could then be transformed as explained above to identify the potentially best performing algorithm.

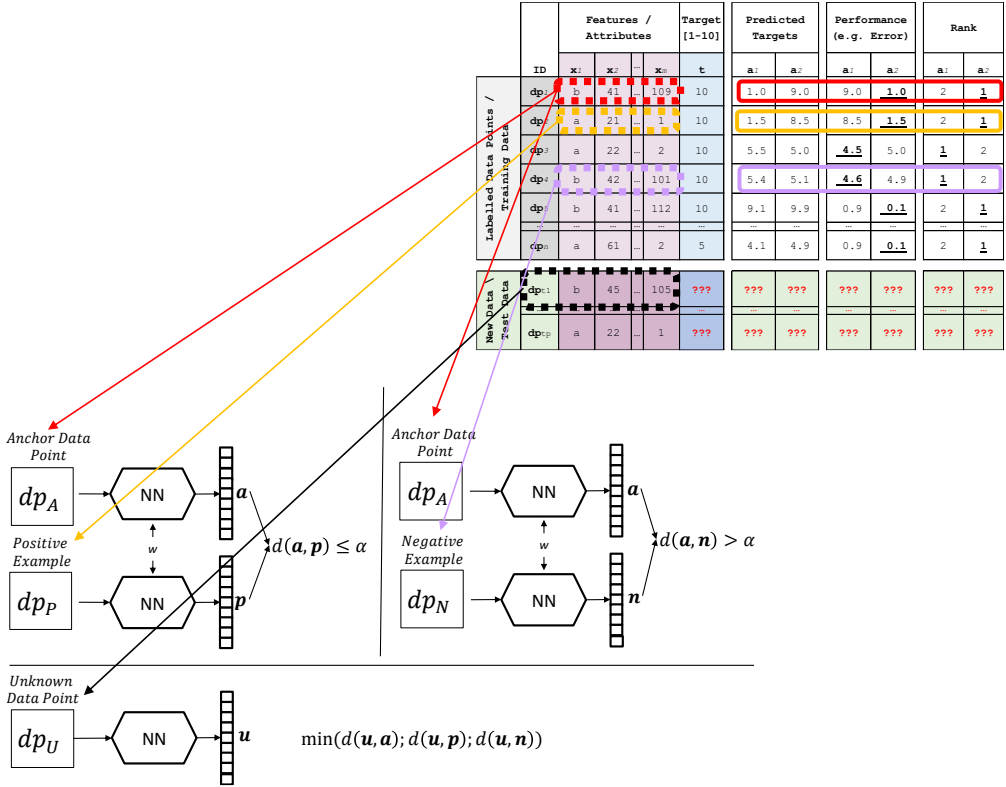

**Figure 9: Illustration of a Siamese Neural Network architecture for algorithm selection**

### 3.2   Algorithm Performance Persona & Algorithm Performance Space

The concept of a person is naturally given in image classification, and hence it is given what images in a dataset can serve as positive and negative samples[2]. Two photos either are of the same class, i.e. they show the same person (positive sample), or they are not (negative sample). For algorithm selection such a concept is not naturally given. Hence, for Siamese Algorithm Selection, an equivalent is needed to a 'person' in image classification. Only with such an equivalent it will be possible to select training pairs optimally.

We term that equivalent an 'Algorithm Performance Persona' (APP). We adopt the term 'Persona' from user experience (UX) design and market research, which define 'Personas' as follows:

---

[2] There are additional caveats to select training samples such as selecting a mix of simple and difficult to learn samples, which we will cover later.

> *"Personas are archetypical users whose goals and characteristics represent the needs of a larger group of users."* [54]

> *"A user persona is a fictional representation of a business's ideal customer; they are generally based on user research and incorporate the needs, goals, and observed behavior patterns of your target audience. User personas are constructed using sample qualitative and quantitative data that is collected from actual users."* [55]

Such a fictitious user, or person, is what we need for Siamese Algorithm Selection, and the characteristics of that fictitious 'person' should relate to the performance of algorithms on the given data points. For algorithm selection, this persona does not necessarily have to be fictitious. It could be a real data point that represents a larger group of similar data points. We therefore define an 'Algorithm Performance Persona' (APP) as follows:

> *"Algorithm Performance Personas (APP) are archetypical data points, fictitious clusters of data points, or small sub-spaces in the performance space whose performance characteristics represent the characteristics of a larger group of data points."*

We illustrate the concept of APPs in Figure 10. The top right hand-side of the illustration shows the instances from our example plotted to the performance space.

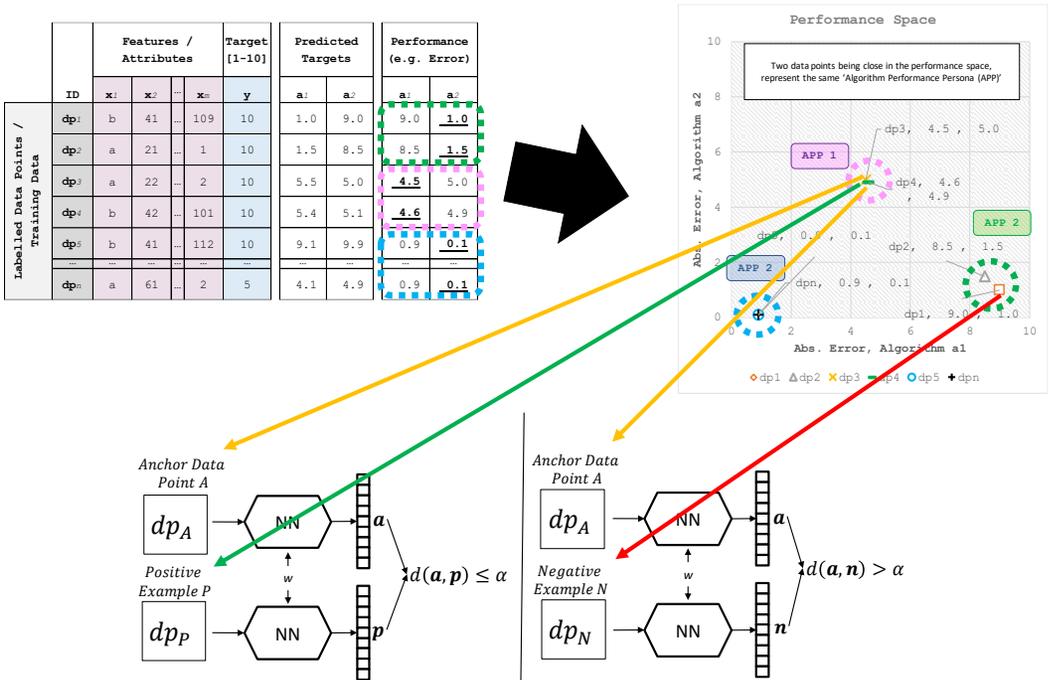

**Figure 10: Illustration of Algorithm Performance Persona**

The performance space has *n* dimensions, each dimension representing the performance of one of the single algorithms. In the example, performance is measured as absolute error (we will later introduce a more suitable metric). Data points close in the performance space would be considered to represent the same Algorithm Performance Persona. Data points belonging to the same APP would then be used as anchor and positive training pair. Datapoints of a different persona would be used as negative training sample. Our example is simple and only has two dimensions. However, in more realistic scenarios, with maybe a dozen of algorithms, the performance space would become a relatively high dimensional space in which thousands of APPs could exist. If our concept was to be extended to algorithm configurations and pipelines, the dimensions could reach even larger numbers.

On first glance, an Algorithm Performance Persona may seem the same like a cluster. However, as we will show later, clustering is not an ideal technique to identify Algorithm Performance Personas. Also, we believe that a rather generic term is needed that is not assigned with a specific technique such as clustering as we are optimistic that in the future many different techniques may be proposed to identify APPs. The approach we will propose is just one of potentially many.

### 3.3 Design of the Performance Space (Performance Metric & Similarity Function)

The key to Siamese Algorithm Selection and identifying Algorithm Performance Personas, and hence suitable training samples, is the performance space. The performance space in turn depends on a) the metric to express performance, and b) the similarity measurement and defining boundaries of APPs.

There are two rather obvious metrics for the performance space: the ranking of algorithms and the actual performance of an algorithm (e.g. MAE, Accuracy, Precision, ...). However, we consider both choices as suboptimal and will demonstrate in the following section why that is. We then present a novel metric. Since we define the performance space as a vector space, Cosine and Euclidian distance are two rather obvious choices to measure similarity.

Throughout this section, we will continue to use the example we used previously. More specifically, we look at the case in which we have chosen $dp_1$ as anchor point (Figure 11), and we aim to identify a positive and negative training pair for the Siamese Network training.

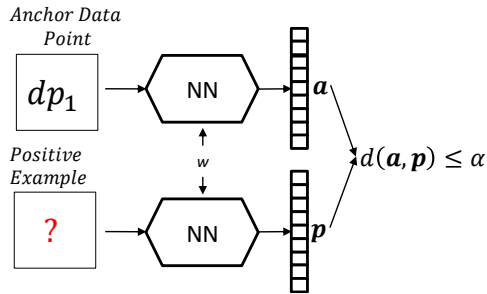

**Figure 11**

In Figure 12 we augmented our previous example by our intuition of what data points should be similar to $dp_1$. As explained previously, $dp_2$ would be highly similar. Data points $dp_5$ and $dp_n$ would be somewhat similar to $dp_1$. 'Somewhat' because the same algorithm ($a_2$) performs best on the data points, and the ratio between $a_1$ and $a_2$ is about the same (the error of $a_1$ is around 9 times as high as for $a_2$ on the instances). On the other hand, the magnitude of the error on $dp_5$ and $dp_n$ is smaller than on $dp_1$ and $dp_2$ (errors of around 0.9 vs 9 and 0.1 vs 1 respectively). $dp_3$ and $dp_4$ are not similar to $dp_1$ as a different algorithm performs best ($a_1$) and the ratio differs (both $a_1$ and $a_2$ perform relatively similar).

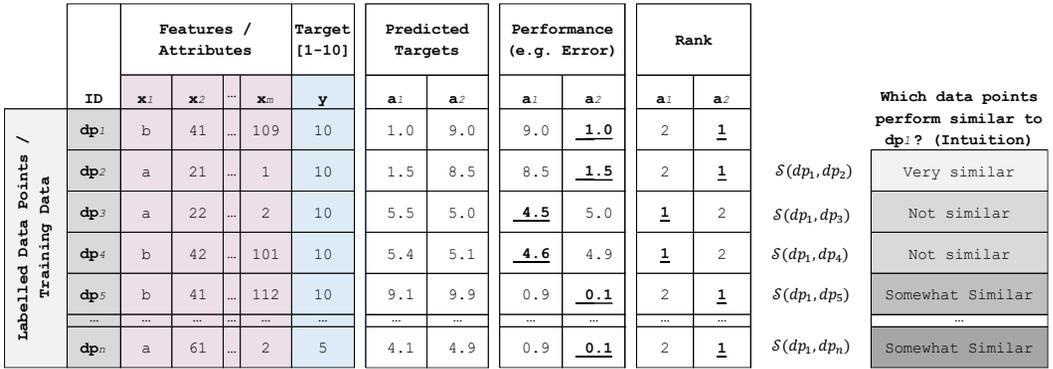

**Figure 12**

*3.3.1 Rank.* On first glance the rank of an algorithm for a data point may seem like a good choice. In that case, two data points for which the same algorithm performs best, second best... worst, would be chosen as positive training pair. Using the rank for defining Algorithm Performance Personas would, probably, not suffice. With rank, there is only a small number of possible APPs for any given set of algorithms. For instance, with 3 algorithms, there were only six possible combinations (1,2,3; 1,3,2; 2,1,3; 2,3,1; 3,2,1; 3,1,2) and hence APPs. Such a low number would likely not be enough for training a Siamese Neural Network which is typically trained on at least dozens of classes.

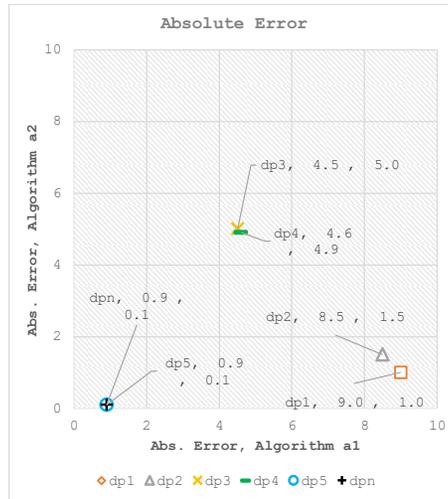

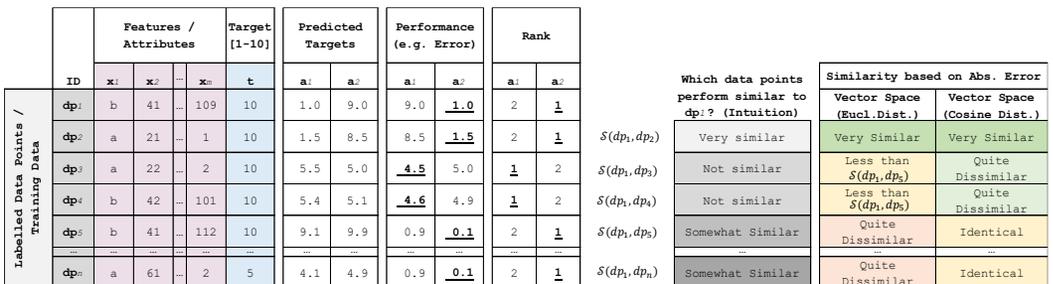

**Figure 13: Performance Space based on Absolute Error**

*3.3.2 Absolute Error.* The next rather obvious choice to design the algorithm performance space would be the 'ordinary' performance metric, e.g. error, accuracy, precision, etc. In our example with two algorithms, the performance space would look like in Figure 13. In such a space, similarity among data points is not fully as we desire (Figure 12). Measured by Euclidian distance, $dp_1$ and $dp_2$ would indeed be similar. However, distance between $dp_1$ and $dp_5$ as well as between $dp_1$ and $dp_n$ would be high, which is not what we want. Also, distance between $dp_1$ and $dp_3$ is lower than between $dp_1$ and $dp_5$, which is not what we want. Cosine distance would lead to more favorable result, yet not exactly what we want. With Cosine distance, for instance, $dp_1$, $dp_5$, and $dp_n$ would be considered identical. As outlined before, we would aim for a metric that defines them as 'somewhat' similar, but not identical.

*3.3.3 RIIPxMPRE (Our Metric).* We propose a novel performance metric (or normalization scheme), named *RIIP_MPRE*: *Relative Intra-Instance-Performance * Max-Possible Relative Error.*

$$RIIP\_MPRE = P_{a_k}(\boldsymbol{x_i}, y_i) = (1 - \varepsilon_{k,i}) * I_{k,i}$$

The *Relative Intra-Instance Performance* (RIIP), or $I_{k,i}$, measures how well an algorithm performs compared to the other algorithms on that instance. For every data point, the best performing algorithms achieves a RIIP of 1 (or 100%). The other algorithms receive values between 0 and 1, indicating how close their original performance is to the best performing one. For instance, $a_1$ has an error of 8.5 on $dp_2$ and $a_2$ has an error of 1.5. This means, $a_1$ only is 18% as good as algorithm $a_2$, and hence RIIP is 0.18 or 18%. RIIP is inspired by other similar metrics for pairwise comparisons and landmarkers, which are relatively commonly used for automated algorithm selection. However, to the best of our knowledge the metrics were used in different contexts, and typically for binary comparisons. Also, RIIP alone is not sufficient for our purpose.

$$I_{k,i} = \frac{P_{a^*}(\boldsymbol{x_i}, y_i)}{|y_i - \hat{y}_{k,i}|}$$

The *Max-Possible Relative Error (MPRE)*, or $\varepsilon_{k,i}$, takes into consideration that for different data points the scale may vary. This is true for many regression problems, particularly in the field of recommender systems. For instance, for movie recommendations, ratings are often made on a scale between 1 and 10). If the actual target rating is 4, then the maximum possible error an algorithm could make is 6 (i.e. when the algorithm predicts 10). However, if the actual rating is 9, then the maximum error an algorithm can make is 8 (i.e. when the predicted rating is 1). In our example, the absolute errors for $dp5$ and $dpn$ are identical (0.9 and 0.1 respectively). However, the target for $dp5$ is 10, and for $dpn$ is 5. Hence, we would argue that the algorithms performed better on $dp5$ than on $dpn$. For $dp5$, the algorithms are only 9% and 1% off the actual target, while for $dpn$ the algorithms are off 18% and 2% respectively. *MPRE* takes this into consideration by expressing an algorithm's performance as the error relative to the maximum possible error. Of course, this metric is only relevant for scenarios where different data points have different maximum possible errors.

$$\varepsilon_{k,i} = \frac{|y_i - \hat{y}_{k,i}|}{\varepsilon_{M,i}}$$

$$\varepsilon_{M,i} = \max(\hat{y}_{k,i} - \mathcal{B}_L, \mathcal{B}_U - \hat{y}_{k,i})$$

| Symbol | Description |
|---|---|
| $P_{a_k}(x_i, y_i)$ | Performance of algorithm $a_k$ on $x_i$ with target $y_i \in [\mathcal{B}_L, \mathcal{B}_U]$ |
| $\varepsilon_{k,i}$ | The error of $a_k$ on $x_i$ relative to the maximal possible error for $x_i$ |
| $I_{k,i}$ | Intra-instance Performance of $a_k$ on $x_i$ relative to the best algorithm $a^*$ on instance $x_i$, i.e. $P_{a^*}(x_i, y_i)$ |
| $P_{a^*}(x_i, y_i)$ | The best performance (lowest error) achieved on $x_i$ with $a^*$ |
| $\varepsilon_{M,i}$ | Maximal possible error for instance $x_i$ |
| $\mathcal{B}_U$ | Upper global bound of the target |
| $\mathcal{B}_L$ | Lower global bound of the target (typically 0 or 1) |
| $\hat{y}_{k,i}$ | Predicted target for $x_i$ by algorithm $a_k$ |
| $y_i$ | Target for $x_i$ |

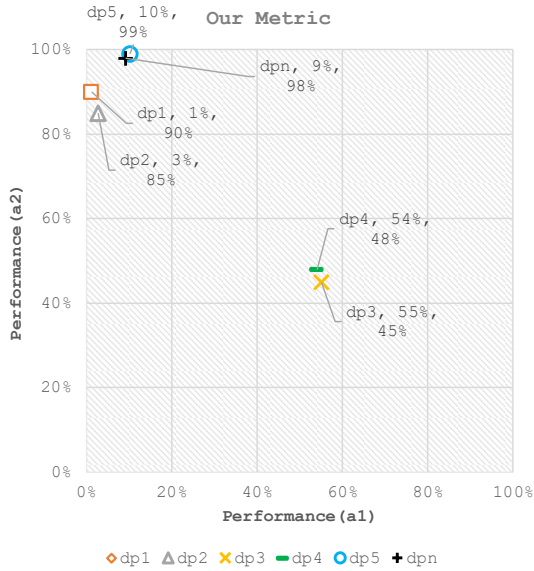

**Figure 14: Performance Space based on our novel metric**

Figure 14 illustrates the performance space with our novel metric. The lower part of Figure 14 also shows how similarity with both Cosine and Euclidian distance aligns with our expectations. At least in our example, our novel metric satisfied our intuition perfectly.

　　　　　We also conducted a small experiment to identify the effect the metric has on the performance space. We run five machine learning algorithms (Lasso, SGD, CatBoost, MLP, Random Forrest) on the Lending Club Loan (LCL) dataset[3]. We picked the loan's interest rate as a target, which varies between near 0 and around 30% and an average of 13.06%. We plotted a sample of 1,000 instances to the performance spaces. For sake of illustration we plot multiple 2-dimensional spaces (Figure 15), while, in practice, there would only be one 5-dimensional space.

---
[3] https://www.kaggle.com/wendykan/lending-club-loan-data

For instance, the very top-left performance space in Figure 15 shows how Random Forrest and MLP perform on the 1,000 instances. All instances are grouped relatively close together. The same applies in all other performance spaces. Selecting appropriate training samples in such a dense space would be difficult and error prone. In contrast, Figure 16 shows the same instances, but with our novel performance metric. The instances are distributed much wider over the performance space. This *should* make this performance space more suitable to select training pairs for the Siamese Algorithm Selection training.

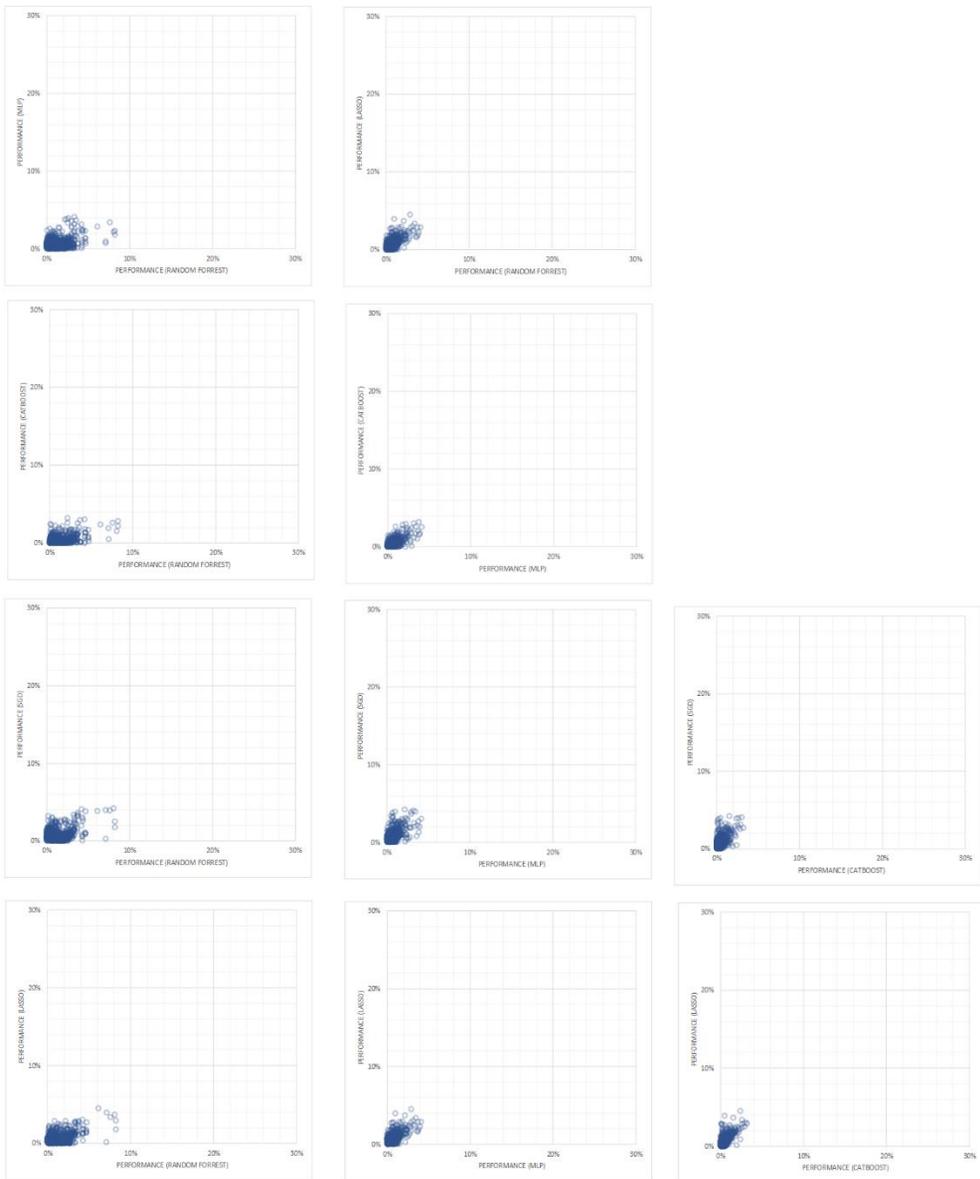

**Figure 15: 1,000 Samples in the performance space (2 Dimensions Each); Mean Absolute Error**

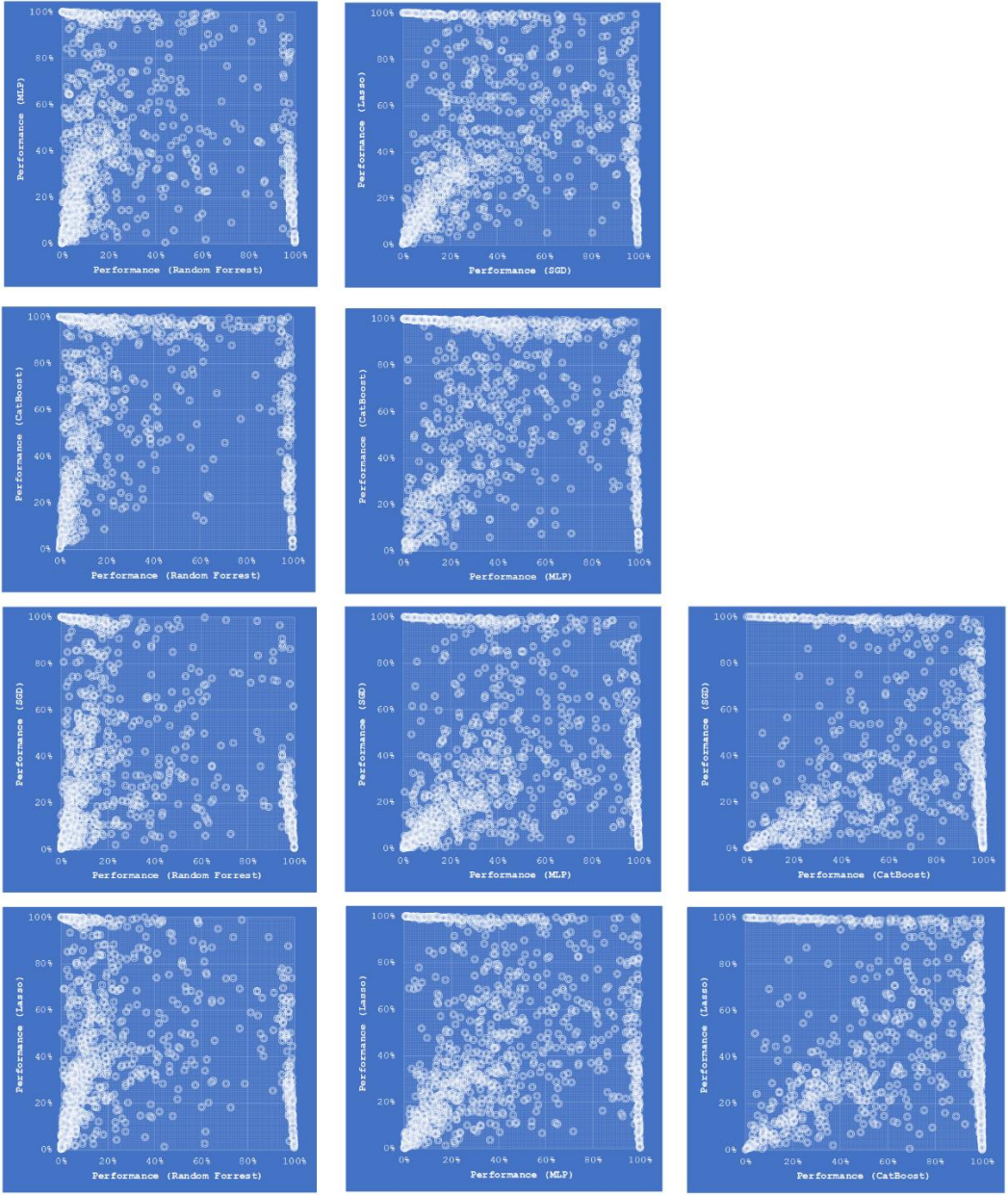

**Figure 16: 1,000 Samples in the performance space (2 Dimensions Each); Our Novel Metric**

## 3.4 Selection of Training Samples (Anchor, Positive, Negative)

*sClustering* may seem as obvious choice to identify Algorithm Performance Personas, and hence training pairs. In that case, data points in the performance space would be clustered, and data points of the same cluster could be used as positive training pairs (Figure 17). Data points from other clusters could be used as negative samples.

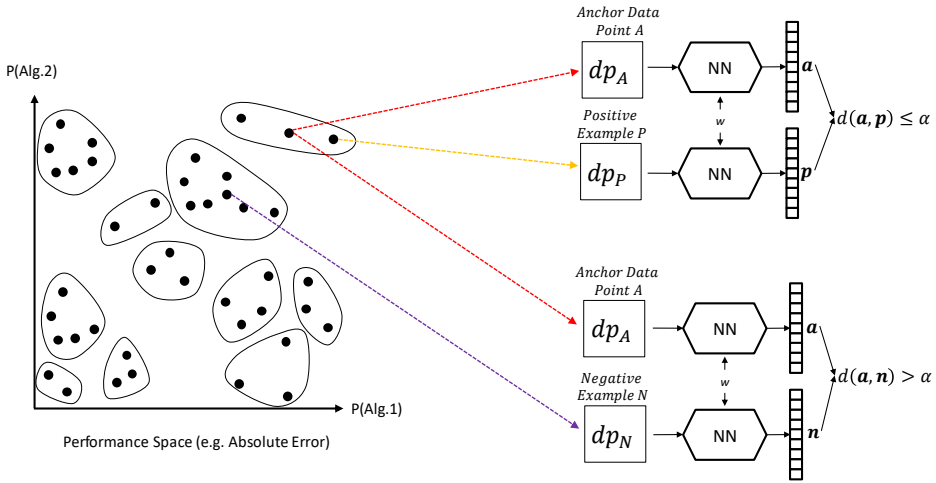

**Figure 17: Clustering in the performance space to identify positive and negative training samples**

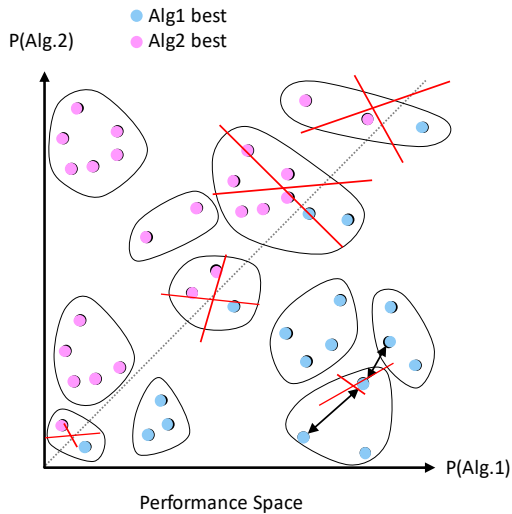

**Figure 18: Potential problems when using clustering**

We would expect though that clustering will not perform optimally for two reasons. First, there likely would be clusters with data points for which different algorithms perform best. This is illustrated in Figure 18 for a 2-dimensional performance space. Algorithm 1 (blue) performs best for all data points below the 45° line. Algorithm 2 (pink) performs best for all instances above the 45° line. A clustering technique likely would build clusters that include datapoints from both below and above the 45° line. This likely would not lead to precise prediction later about which algorithms will perform best for a new data point. Second, the distance between two data points in the same cluster might be larger than the distance between data points in different clusters. Treating data points in different clusters but within close distance as different APPs is counter-intuitive.

## ACKNOWLEDGEMENTS

This research was partly conducted with the financial support of the ADAPT SFI Research Centre at Trinity College Dublin. The ADAPT SFI Centre for Digital Media Technology is funded by Science Foundation Ireland through the SFI Research Centres Programme and is co-funded under the European Regional Development Fund (ERDF) through Grant #13/RC/2106.

Siamese Meta-Learning and Algorithm Selection [Proposal]	39:19